\documentclass[10pt]{article}

\usepackage[a4paper,textwidth=18cm,textheight=24cm,top=2.85cm, bottom=2.85cm, left=1.5cm, right=1.5cm]{geometry}

\usepackage{main}

\makeatletter
\long\def\@makecaption#1#2{%
\vskip\abovecaptionskip
\sbox\@tempboxa{#1. #2}%
\ifdim \wd\@tempboxa >\hsize
#1. #2\par
\else
\global \@minipagefalse
\hb@xt@\hsize{\box\@tempboxa\hfil}%
\fi
\vskip\belowcaptionskip}
\makeatother

\usepackage{lmodern}
\usepackage{multirow}
\usepackage{graphicx}
\usepackage{times}
\usepackage{algorithm}
\usepackage{algorithmic}

\begin{document}
\noindent

\bibliographystyle{ieeetr}

\title{Latent Vector Expansion using Autoencoder for Anomaly Detection}

\authorname{UJu Gim\thanks{Correspondence author: gim.uju1217@sk.com} , YeongHyeon Park}
\authoraddr{SK Planet Co., Ltd.}
\maketitle

\abstract
Deep learning methods can classify various unstructured data such as images, language, and voice as input data.
As the task of classifying anomalies becomes more important in the real world, various methods exist for classifying using deep learning with data collected in the real world. 
As the task of classifying anomalies becomes more important in the real world, there are various methods for classifying using deep learning with data collected in the real world. Among the various methods, the representative approach is a method of extracting and learning the main features based on a transition model from pre-trained models, and a method of learning an autoencoder-based structure only with normal data and classifying it as abnormal through a threshold value. 
However, if the dataset is imbalanced, even the state-of-the-arts models do not achieve good performance. This can be addressed by augmenting normal and abnormal features in imbalanced data as features with strong distinction. We use the features of the autoencoder to 
train latent vectors from low to high dimensionality. We train normal and abnormal data as a feature that has a strong distinction among the features of imbalanced data.
We propose a latent vector expansion autoencoder model that improves classification performance at imbalanced data. The proposed method shows performance improvement compared to the basic autoencoder using imbalanced anomaly dataset.
\section{Introduction}
\label{sec:introduction}
Nowadays, using deep learning through data sets collected in the real world, image classification \cite{lecun1998gradient, krizhevsky2009learning} and abnormal situation determination \cite{park2020anomaly, sultani2018real} among various unstructured data are in progress.
In order to classify images and determine abnormal situations, data is collected and meaningful feature values are extracted from the collected datasets. The extracted significant feature values enable the machine to recognize the input data of the data set. 
However, there is a problem that the machine does not recognize the difference between normal and abnormal even if data pre-processing is performed among the collected data, and it is difficult to compare the differences between classes.
To overcome this limitation, we train an autoencoder with a simple change in the structure of the basic autoencoder \cite{hinton2006fast}. Normal and abnormal data are augmented through the latent vector generated by the encoder from trained autoencoder. In order to extract strong feature values for normal and abnormal classification of the generated latent vectors, the low dimensionality is extended to the high dimensionality.
This improves classification performance by strongly demarcating the decision boundaries of each class. We show an increase in performance over the basic autoencoder by experimenting with imbalanced anomaly dataset.
\section{Related work}
\label{sec:related_work}
The basic autoencoder consists of an encoder and a decoder. By reducing the dimension of the input value through the encoder, the value is compressed into features representative of the input value. After that, the reduced-dimensional latent vector is put as the input value of the decoder and the vector is restored to generate a value similar to the input value of the encoder. The purpose of a basic autoencoder is to train meaningful data from a reduced-dimensional latent vector in the encoder.
Variational autoencoder, similar in structure to autoencoder, expresses probability distribution in input data through latent vectors \cite{kingma2013auto}. After that, data is generated through the decoder. The difference between an autoencoder and a variational autoencoder is as follows. The autoencoder recovers the reduced data by reducing the dimensions of the data. Conversely, a variational autoencoder is a generative model that generates similar data in a probability distribution. The reason we use the autoencoder is that it is used only for the purpose of reducing the dimension, not for the purpose of data generation.

Contrary to the encoder of the autoencoder, the kernel trick is to extend the characteristics of data from low to high dimensionality to distinguish the boundary determination of each class \cite{scholkopf2001kernel}. Support vector machine (SVM) \cite{noble2006support} improves the classification performance of input data from low-dimensional linear models to high-dimensional models. The kernel trick does not actually expand the data, but rather computes the scalar product of the data through the extended properties. We solve the classification problem by explosively expanding the dimensionality of the latent vector of the autoencoder, which is difficult to classify due to the reduced dimensionality of feature values in low dimensions.
\begin{figure}[ht]
    \begin{center}
		\includegraphics[width=1\linewidth]{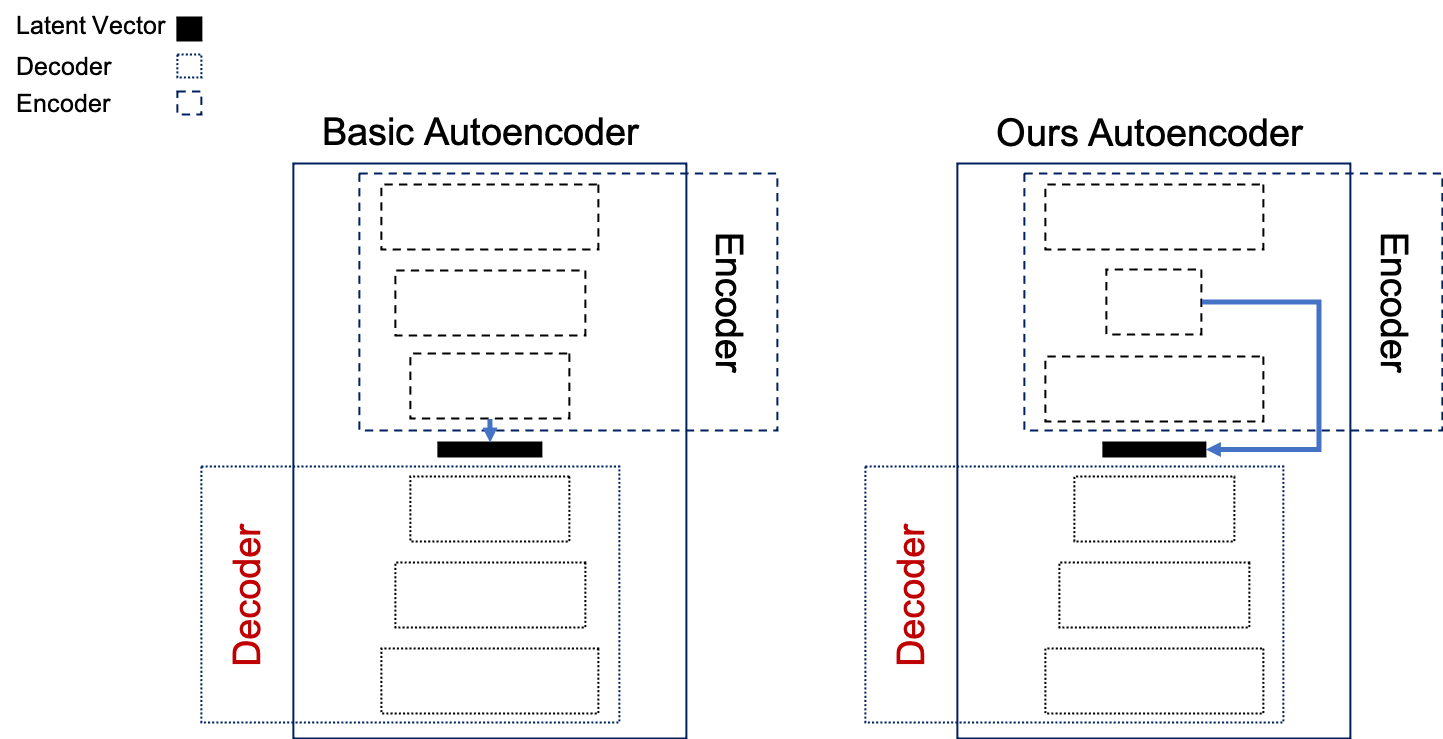}
	\end{center}
	\caption{Comparison of autoencoder structure}
	\label{fig:fig1}
\end{figure}
\section{Latent vector expansion-based autoencoder}
\label{sec:proposed_approach}
In this section, we present the latent vector extension-based autoencoder model, which is the method proposed in this paper. The Basic Autoencoder (BA) in \ref{fig:ae-structure} shows the basic autoencoder and our autoencoder, which is restored through the decoder after reducing the dimension of the encoder. Each paragraph describes our autoencoder and Latent Vector Expansion Network. We present our autoencoders different from BA. 

First, we add the first linear layer to the encoder of the autoencoder, which specifies the size of the input value data. The second layer is set as a linear layer that reduces in proportion to the size of the first layer, and the input and output values are the same. Add the ReLu activation function to the first and second layers. The third layer of the encoder sets the output value of the second layer as an input value and sets the input value of the first layer as an output value. Subsequently, the decoder reverses the parameters of the set layer of the encoder, and then sets the activation function of the last layer to sigmoid. We adopted ours, which expands the output of the second layer based on the output of the third layer through experiments. Table \ref{tab:AE-comparison} shows the outstanding performance of ours, which expands the output value in the third layer after reducing the size of each input data compared to the latent vector generated through the BA structure. A description of the linear model is given in the next paragraph, Latent Vector Expansion Network.
\begin{table}[ht]
    \centering
    \caption{Autoencoder comparison with AUROC.}
    \begin{tabular}{l|c}
        \hline
            \textbf{Method} & \textbf{AUROC} \\       
        \hline
            Linear model w/ BA & 0.872 \\
            Linear model w/ Ours & \textbf{0.951} \\
        \hline
    \end{tabular}
    \label{tab:AE-comparison}
\end{table}

Second, we construct a Latent Vector Expansion network that trains latent vectors from the trained autoencoder's encoder. This network simply consists of two linear layers and a linear model as an output. After setting the input value of the first layer as the output value of the encoder, the output value expands the dimension to 1,024. After that, the activation function uses ReLU and the dropout is set to 0.5. In addition, a linear layer is set to set the output to 1 for normal and abnormal binary classification, and the final value is output through a log sigmoid. We show the 1024-dimensional performance through Table 2. A detailed description of each table is provided in Chapter 4.
\begin{table}[ht]
    \centering
    \caption{Expansion dimension comparison with AUROC.}
    \begin{tabular}{c|cc}
        \hline
            \textbf{Method} & \textbf{Expansion dimension} & \textbf{AUROC} \\      
        \hline
        \multirow{4}{*}{Ours w/ expansion} & 128 & 0.969 \\  
             & 256 & 0.968 \\  
             & 512 & 0.969 \\  
             & \textbf{1,024} & \textbf{0.970} \\  
         \hline

    \end{tabular}
    \label{tab:expansion-comparison}
\end{table}
\section{Experiments}
\label{sec:experiments}
In this chapter, we present the dataset and preprocessing used for the experiment of our proposed latent vector expansion-based autoencoder. After that, the parameter settings used in each of Tables 1, 2 and 3 will be described. The experiment used credit card data set, and the main purpose of this data set is to classify fraud. It is an imbalanced data set with 492 abnormal out of a total of 284,807 data, accounting for only about 0.17\% \cite{le2004machine}. We performed Min-Max scaling to prevent overfitting of imbalanced dataset, and conduct experiments through K-Fold, which is cross validation technic. For all models used in the experiment, the epoch is 20, the learning rate is 0.001, the optimizer used Adam, and the loss used binary cross entropy. The difference is that the autoencoder uses Mean Square error for Epoch 50 and loss during training.

First in Table \ref{tab:AE-comparison}, w/ BA and w/ Ours set the encoder latent vector as the input value of the linear model after learning the autoencoder.

Table \ref{tab:expansion-comparison} conducts the experiment by changing the extension dimension of the first layer of the latent vector extension network.

Table \ref{tab:Linearmodel} shows that each model conducts an experiment through the input value of the K-fold training dataset without an autoencoder. A latent vector extension network is used as a linear model. The w/o expansion changed the expansion dimension of the first layer of the linear model to 10, and the w/ expansion changed the expansion dimension to 1,024 and conducted the experiment.

\subsection{Experiment result}
In this chapter, we explain the reasons for the performance difference between Table 3 and Table 4. We show the BA model and our model performance using K-Fold in Table 4. Compared with Table 3, Table 4 has relatively not good performance compared to the linear models, but the input value of the linear model is basic data without reducing the dimension of the data. Since our model is data extracted from the encoder of a trained autoencoder, there are two advantages. The first is that it can be lightweight when learning the model, and the second can be faster than the basic data when inference is performed. In the comparison of models in Table 4, it shows superior performance in all Folds compared to when the BA is used. For this reason, our proposed Latent vector expansion-based autoencoder and latent vector expansion method serve to improve performance.

Figure 2 shows the PCA analysis of 2 and 2 folds showing the best performance and 5 and 7 folds showing not good performance in the two models (left is ours, right is BA in Figure 2)  in Table 4. It can be seen that models classifier even the data set with not good data distribution using an expansion latent vector.

\begin{table}[ht]
    \centering
    \caption{Linear model Comparison of with out expansion and with expansion models.}
    \begin{tabular}{c|c|c}
        \hline
        \multirow{3}{*}{\textbf{Fold}}  
        & \textbf{Linear model} & \textbf{Linear model}\\
        & \textbf{w/o expansion} &  \textbf{w/ expansion} \\
        & \textbf{AUROC} & \textbf{AUROC}\\
        \hline
            1 & \textbf{0.994} & 0.990 \\
            2 & \textbf{0.995} & \textbf{0.995} \\
            3 & 0.998 & \textbf{0.999} \\
            4 & \textbf{0.935} & 0.927 \\
            5 & 0.964 & \textbf{0.972} \\
            6 & 0.986 & \textbf{0.987} \\
            7 & 0.990 & \textbf{0.991} \\
            8 & 0.970 & \textbf{0.978} \\
            9 & 0.983 & \textbf{0.985} \\
            10 & 0.973 & \textbf{0.981} \\
        \hline
    \end{tabular}
    \label{tab:Linearmodel}
\end{table}

\section{Conclusion}
\label{sec:conclusion}
In this paper, we proposed a latent vector extension-based autoencoder model that can show that our methodology is superior to that of the basic autoencoder structure and encoder for abnormal situation data. Through experiments, the proposed model showed that the abnormal situation detection performance was superior to the existing model through the credit card dataset. We will future experiment with images and various data sets.

\begin{table}[ht]
    \centering
    \caption{BA model and our model performance comparison.}
    \begin{tabular}{c|c|c}
        \hline
        \multirow{3}{*}{\textbf{Fold}}  
        & \textbf{BA} & \textbf{Ours}\\
        & \textbf{w/ expansion} & \textbf{w/ expansion} \\
        & \textbf{AUROC} & \textbf{AUROC}\\
        \hline
        1 & 0.939 &\textbf{0.957} \\
        2 & 0.943 &\textbf{0.969} \\
        3 & 0.884& \textbf{0.930} \\
        4 & 0.851&\textbf{0.901} \\
        5 & 0.864&\textbf{0.898} \\
        6 & 0.919&\textbf{0.945} \\
        7 & 0.790&\textbf{0.920} \\
        8 & 0.923&\textbf{0.954} \\
        9 & 0.884&\textbf{0.910} \\
        10 & 0.847&\textbf{0.935} \\
        \hline
    \end{tabular}
    \label{tab:oursmodel}
\end{table}

\begin{figure}[ht]
    \begin{center}
		\includegraphics[width=1\linewidth]{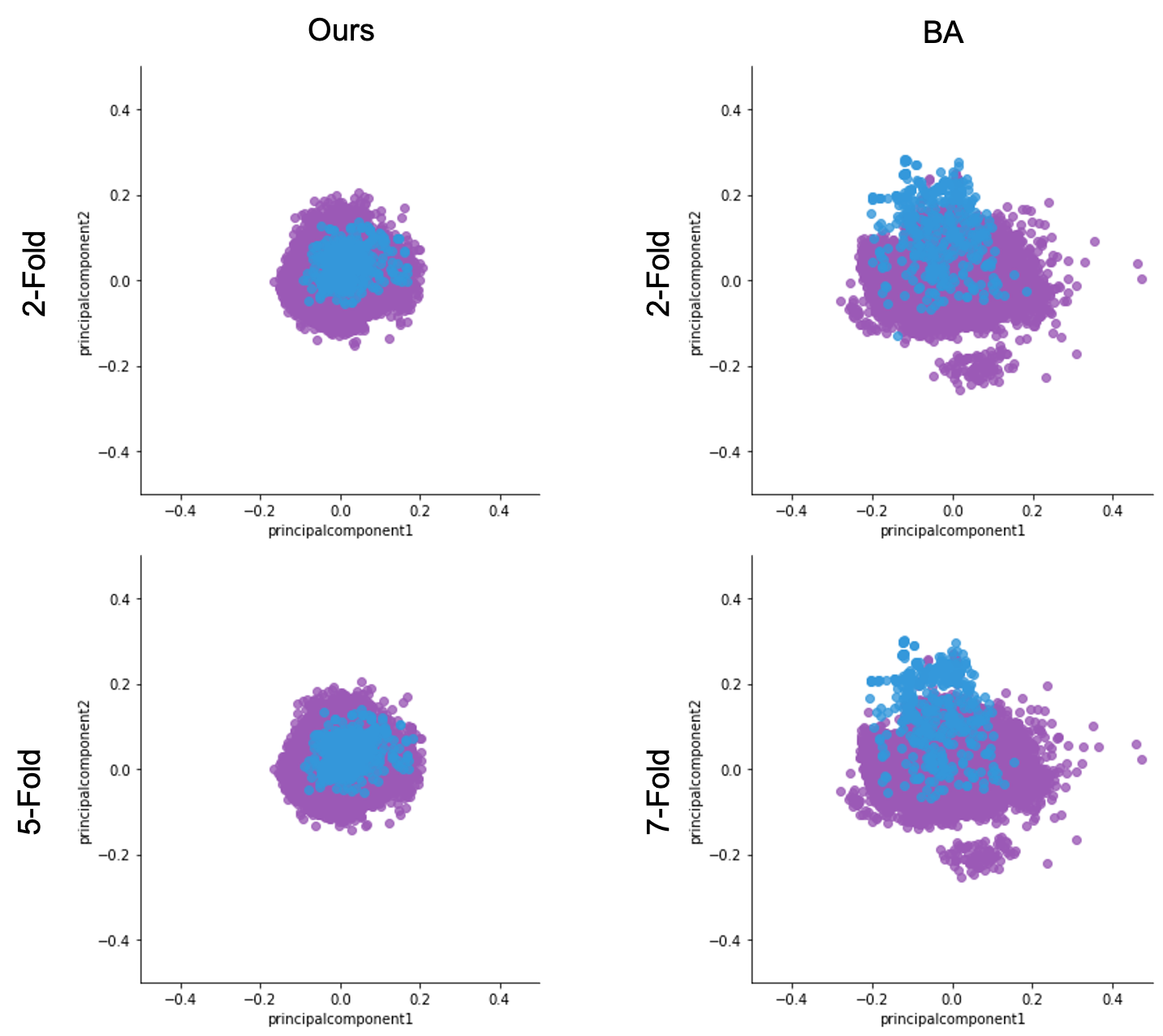}
	\end{center}
	\caption{BA model and our model PCA Comparison}
	\label{fig-fig2}
\end{figure}


\bibliography{main.bib} 

\end{document}